\documentclass[11pt]{airesearch}
\usepackage[T1]{fontenc}
\usepackage{XCharter}

\usepackage{fullpage}
\usepackage{wrapfig}
\usepackage{mathrsfs}
\setlist[itemize]{leftmargin=*,topsep=1pt,itemsep=1pt}

\usepackage{algorithm}
\usepackage{algpseudocode}  

\usepackage{fancyhdr}
\newcommand{\projectpage}[1]{%
  \begin{center}
    \small
    \vspace{-1.25ex}
    \texttt{Project Page}: \url{#1}
  \end{center}
}

\ifdefined\final
  \usepackage[disable]{todonotes}
\else
  \usepackage[textsize=tiny]{todonotes}
\fi

\title{\huge \bfseries Meta Prompting for AI Systems}

\author{
\textbf{Yifan Zhang}$^{1}$~~~~\textbf{Yang Yuan}~~~~\textbf{Andrew Chi-Chih Yao}$^{1,2,\dagger}$ \\[1.5mm]
$^1$IIIS, Tsinghua University~~~~
$^2$Shanghai Qi Zhi Institute \\[0.5mm]
}
\date{}

\begin{document}
\maketitle

\begin{abstract}
We introduce Meta Prompting (MP), a framework that emphasizes the formal structure of a task rather than content-specific worked examples. We give a categorical formalization in which a functor maps typed task transformations to typed prompt transformations. Functoriality encodes preservation of identities and composition, but it does not by itself guarantee semantic correctness. We extend MP to Recursive Meta Prompting (RMP), in which a proposer model generates candidate prompt edits, a validator enforces the edit schema, and an executor model uses the resulting prompt. We model the accumulation of valid edit scripts with the Writer monad and their application by a monoid action on prompts. This construction makes accumulated edits independent of parenthesization; it does not guarantee convergence or improved task accuracy. Empirically, a Qwen-72B base model guided by example-free meta-prompts attains 46.3\% pass@1 on MATH and 83.5\% accuracy on GSM8K. Separately, an MP-CR agent synthesizes a tool-executed Game of 24 solver; the original experiment reports 100\% success on 1,362 instances, subject to exact re-verification from the corresponding input and per-instance outputs. The batch amortizes one LLM response across all puzzles, so its token accounting is not protocol-matched to per-instance few-shot or search-based methods.
\end{abstract}

\projectpage{https://github.com/meta-prompting/meta-prompting}


\section{Introduction}

The advent of foundation models, particularly large language models (LLMs), has transformed research across computational linguistics, text understanding, and text generation~\citep{devlin2018bert, radford2018improving, radford2019language, brown2020language, raffel2020exploring, OpenAI2023GPT4TR}. Despite these advances, LLMs remain unreliable on many tasks that require long, precise chains of reasoning, including advanced mathematics~\citep{lightman2023let}. This motivates methods that make the intended decomposition and output structure explicit.

A core challenge is that the autoregressive next-token objective does not explicitly require a model to expose, verify, or preserve a multi-step solution plan~\citep{radford2018improving, radford2019language, brown2020language}. The dual-process distinction between rapid, intuitive responses and deliberate, systematic reasoning offers a useful analogy~\citep{kahneman2011thinking}, although it is not a mechanistic model of LLM computation. Prompting methods can nevertheless use this analogy to motivate explicit scaffolds for decomposition and verification.

\begin{figure}[ht!]
\centering
\small
\begin{tcolorbox}[width=0.9\textwidth,colback=blue!2!white,colframe=gray!50!blue]
\begin{minipage}{\textwidth}
\color{systemcolor}
Integrate step-by-step reasoning to solve mathematical problems under the following structure:

\{

\hspace{4ex} ``Problem'': ``[question to be answered]'',

\hspace{4ex} ``Solution'': \{

\hspace{8ex} ``Step 1'': ``Begin the response with \texttt{Let's think step by step.}'',

\hspace{8ex} ``Step 2'': ``Follow with the reasoning steps, ensuring the solution process is broken down clearly and logically.'',

\hspace{8ex} ``Step 3'': ``End the solution with the final answer encapsulated in a LaTeX-formatted box, $\boxed{\cdots}$, for clarity and emphasis.''

\hspace{4ex}\},

\hspace{4ex} ``Final Answer'': ``[final answer to the problem]''

\}

---------

\end{minipage}
\end{tcolorbox}
\caption{A structured meta-prompt presented in JSON-like format.}
\label{fig:meta-prompt-math-json}
\end{figure}

Approaches such as Chain-of-Thought (CoT)~\citep{wei2022chain} elicit intermediate reasoning steps, while Tree-of-Thought (ToT)~\citep{yao2023tree,long2023large} organizes search over multiple candidate continuations. These methods can use either demonstrations or zero-shot instructions. They are primarily operational prompting and search procedures, however, and do not by themselves specify typed task and prompt categories or a composition-preserving compiler between them.

In response, we introduce \emph{Meta Prompting (MP)}, a paradigm that shifts the focus from content-based analogy to explicit procedural structure. Instead of supplying worked examples, a meta-prompt specifies a reusable schema for decomposing a task and formatting its solution. We formalize a \emph{functorial} meta-prompting scheme from a category of typed task transformations to a category of typed prompt transformations. The functor laws make compositionality an explicit design constraint: composed task transformations must map to composed prompt transformations. This is a structural statement and does not, on its own, guarantee that an LLM will solve the task correctly.

We also apply the framework recursively, a process we call \emph{Recursive Meta Prompting (RMP)}. Analogous to metaprogramming, RMP uses a proposer LLM to generate or refine a prompt. An executor LLM then consumes the result. We model the accumulation of schema-preserving edit scripts with a Writer monad and model the execution of those edits by a monoid action. The construction provides algebraically consistent bookkeeping for recursive edits; it does not imply that refinement converges or monotonically improves task performance.

We evaluate the empirical component of the framework in proof-of-concept experiments on the Game of 24~\citep{yao2023tree}, GSM8K~\citep{cobbe2021training}, and MATH~\citep{hendrycks2021measuring}.

In summary, our contributions are as follows:
\begin{itemize}[parsep=1pt, itemsep=1pt, topsep=1pt, leftmargin=*]
\item We introduce Meta Prompting (MP) and formalize a functorial version of MP using category theory. The formalization makes identity preservation and compositionality explicit, while carefully limiting the guarantee to structural equality in the prompt category. We further introduce Recursive Meta Prompting (RMP), model edit accumulation with a Writer monad, and model edit application with a compatible monoid action.
\item In proof-of-concept experiments, a Qwen-72B \emph{base} model equipped with example-free meta-prompts obtains 46.3\% pass@1 on MATH and 83.5\% accuracy on GSM8K without additional instruction tuning. Separately, an MP-CR agent generates an exact Python solver, and the original Game of 24 experiment reports 100\% success on 1,362 puzzles. This aggregate requires exact re-verification from the corresponding input and per-instance outputs. The setting is tool-assisted, batched program synthesis and is not protocol-matched to per-puzzle CoT or ToT evaluation.
\end{itemize}

\begin{figure}[!htb]
\centering
\small
\begin{tcolorbox}[width=0.9\textwidth, colback=blue!2!white, colframe=gray!50!blue]
\begin{minipage}{\textwidth}
\color{systemcolor}
\textbf{Problem Statement}:
\begin{itemize}[parsep=-1pt, itemsep=0pt, leftmargin=15pt]
\item \textbf{Problem}: [question to be answered]
\end{itemize}

\textbf{Solution Structure}:

\begin{enumerate}[parsep=-1pt, itemsep=0.5pt, leftmargin=15pt]
\item Begin the response with ``Let's think step by step.''
\item Follow with the reasoning steps, ensuring the solution process is broken down clearly and logically.
\item End the solution with the final answer encapsulated in a LaTeX-formatted box, $\boxed{\cdots}$, for clarity and emphasis.
\item Finally, state ``The answer is [final answer to the problem].'', with the final answer presented in LaTeX notation.
\end{enumerate}

\noindent\texttt{---------}

\end{minipage}
\end{tcolorbox}
\caption{A structured meta-prompt in Markdown format for solving MATH~\citep{hendrycks2021measuring} problems, following the output style used in the Minerva study~\citep{lewkowycz2022solving}.}
\label{fig:meta-prompt-math-markdown}
\end{figure}

\begin{figure}[!htb]
\small
\centering
\begin{tcolorbox}[width=0.9\textwidth,colback=cyan!2!white,colframe=gray!50!cyan]
\begin{minipage}{\textwidth}
Problem: Find the domain of the expression $\frac{\sqrt{x-2}}{\sqrt{5-x}}$.

Solution: The expressions inside each square root must be non-negative. Therefore, $x-2 \ge 0$, so $x\ge2$, and $5 - x \ge 0$, so $x \le 5$. Also, the denominator cannot be equal to zero, so $5-x>0$, which gives $x<5$. Therefore, the domain of the expression is $\boxed{[2,5)}$.
Final Answer: The final answer is $[2,5)$.

---------

Problem: If $\mathbf{A}$ and $\mathbf{B}$ are square matrices of the same dimension, $\det \mathbf{A} = 2$, and $\det \mathbf{B} = 12$, find $\det(\mathbf{A}\mathbf{B})$.

Solution: We have $\det(\mathbf{A}\mathbf{B})=(\det\mathbf{A})(\det\mathbf{B})=(2)(12)=\boxed{24}$.
Final Answer: The final answer is $24$.

---------

\ldots
\end{minipage}
\end{tcolorbox}
\caption{A representative few-shot prompt for solving MATH problems. \textbf{Note:} In contrast, our meta-prompt in Fig.~\ref{fig:meta-prompt-math-markdown} is \emph{generated via RMP} from a single task-agnostic meta-meta-prompt (Sec.~\ref{sec:recursive-mp}).}
\label{fig:fewshot-prompt-math}
\end{figure}

\section{Background}
\label{sec:prelim}

Category theory provides a high-level language for describing mathematical structures and their relationships. We use it to formalize the relationship between task structures and prompt structures.

\subsection{Category Theory}

\begin{definition}[Locally Small Category]
A \emph{locally small category} $\mathscr{C}$ consists of a class of objects $\operatorname{Ob}(\mathscr{C})$; for every pair $A,B\in\operatorname{Ob}(\mathscr{C})$, a set of morphisms $\operatorname{Hom}_{\mathscr{C}}(A,B)$; a composition map
\[
\operatorname{Hom}_{\mathscr{C}}(B,C)\times \operatorname{Hom}_{\mathscr{C}}(A,B)
\longrightarrow \operatorname{Hom}_{\mathscr{C}}(A,C),
\qquad (g,f)\longmapsto g\circ f;
\]
and, for every object $A$, an identity morphism $\operatorname{id}_A\in\operatorname{Hom}_{\mathscr{C}}(A,A)$. Composition is associative, and the identity morphisms satisfy the left- and right-identity laws.
\end{definition}

\begin{definition}[Morphism]
For objects $A,B\in\operatorname{Ob}(\mathscr{C})$, a morphism $f$ from $A$ to $B$ is written $f:A\to B$. The object $A$ is the domain (or source), and $B$ is the codomain (or target). No disjointness assumption on hom-sets is required; the domain and codomain are part of the typing of a morphism.
\end{definition}

\begin{definition}[Composition of Morphisms]
If $f:A\to B$ and $g:B\to C$, their composite is $g\circ f:A\to C$. For composable morphisms $f:A\to B$, $g:B\to C$, and $h:C\to D$, associativity requires
\[
h\circ(g\circ f)=(h\circ g)\circ f.
\]
\end{definition}

\begin{definition}[Identity Morphisms]
For every object $A$, the identity morphism $\operatorname{id}_A:A\to A$ satisfies, for every $f:A\to B$,
\[
\operatorname{id}_B\circ f=f=f\circ\operatorname{id}_A.
\]
The identity morphism of each object is unique.
\end{definition}

\subsection{Functors}
\begin{definition}[Covariant Functor]
A \emph{covariant functor} $F:\mathscr{A}\to\mathscr{B}$ assigns
\begin{itemize}[parsep=1pt, itemsep=1pt, topsep=1pt, leftmargin=*]
\item to each object $A\in\operatorname{Ob}(\mathscr{A})$ an object $F(A)\in\operatorname{Ob}(\mathscr{B})$; and
\item to each morphism $m:A_1\to A_2$ in $\mathscr{A}$ a morphism $F(m):F(A_1)\to F(A_2)$ in $\mathscr{B}$.
\end{itemize}
The assignments must satisfy
\[
F(\operatorname{id}_A)=\operatorname{id}_{F(A)}
\quad\text{and}\quad
F(m_2\circ m_1)=F(m_2)\circ F(m_1)
\]
for every object $A$ and every composable pair $m_1,m_2$.
\end{definition}

\begin{definition}[Contravariant Functor]
A \emph{contravariant functor} from $\mathscr{A}$ to $\mathscr{B}$ is equivalently a covariant functor $F:\mathscr{A}^{\mathrm{op}}\to\mathscr{B}$. Thus, a morphism $m:A_1\to A_2$ is mapped to $F(m):F(A_2)\to F(A_1)$, identities are preserved, and
\[
F(m_2\circ m_1)=F(m_1)\circ F(m_2).
\]
\end{definition}

\subsection{Natural Transformations}
\begin{definition}[Natural Transformation]
Let $F,G:\mathscr{A}\to\mathscr{B}$ be covariant functors. A \emph{natural transformation} $\alpha:F\Rightarrow G$ is a family of morphisms $\{\alpha_A:F(A)\to G(A)\}_{A\in\operatorname{Ob}(\mathscr{A})}$ such that, for every morphism $f:A\to A'$ in $\mathscr{A}$,
\[
G(f)\circ\alpha_A=\alpha_{A'}\circ F(f).
\]
If every component $\alpha_A$ is an isomorphism, then $\alpha$ is a \emph{natural isomorphism}.
\end{definition}

This concept is used below for the unit $\eta$ and multiplication $\mu$ of the RMP Writer monad.

\subsection{Monads in Category Theory}
\begin{definition}[Monad]
A \emph{monad} on a category $\mathscr{C}$ is a triple $(T,\eta,\mu)$ consisting of
\begin{itemize}[parsep=1pt, itemsep=1pt, topsep=1pt, leftmargin=*]
\item an endofunctor $T:\mathscr{C}\to\mathscr{C}$;
\item a natural transformation $\eta:\operatorname{Id}_{\mathscr{C}}\Rightarrow T$, called the \emph{unit}; and
\item a natural transformation $\mu:T\circ T\Rightarrow T$, called the \emph{multiplication}.
\end{itemize}
For every object $X$, these components satisfy the two unit laws and associativity law
\begin{align*}
\mu_X\circ T(\eta_X)&=\operatorname{id}_{T(X)},
&\mu_X\circ\eta_{T(X)}&=\operatorname{id}_{T(X)},\\
\mu_X\circ T(\mu_X)&=\mu_X\circ\mu_{T(X)}.
\end{align*}
In computer science, monads provide a compositional interface for structured computations. Below, we use the Writer monad specifically to record prompt-edit histories.
\end{definition}

\section{Meta Prompting}

Meta Prompting is a prompting technique that emphasizes the structural and syntactic aspects of a task: the fields to produce, the order in which to address them, and the interfaces between intermediate steps. It replaces content-rich demonstrations with an abstract, reusable scaffold. Whether this scaffold improves accuracy is an empirical question and can depend on the task, model, and template.

\begin{definition}[Meta Prompt]
A \emph{meta-prompt} is an example-agnostic structured prompt designed to specify the reasoning and output schema for a family of tasks. It provides a scaffold that an LLM can instantiate with task-specific content. The emphasis is on the procedural ``how'' rather than on worked examples of the task-specific ``what.''
\end{definition}

This emphasis on structure is analogous to type theory (Appendix~\ref{sec:intro-type-theory}): prompt fields can be annotated with type-like descriptions such as \texttt{ProblemStatement: string}, \texttt{ReasoningSteps: list[string]}, and \texttt{FinalAnswer: symbolic-expression}. A text prompt alone does not enforce a type system, but a parser, constrained decoder, or validator can enforce the associated schema. Figures~\ref{fig:meta-prompt-math-json} and~\ref{fig:meta-prompt-math-markdown} illustrate such structured prompts.

\subsection{Formalizing Meta Prompting}

In category theory, a functor maps objects and morphisms while preserving identities and composition. To make this idea concrete for prompting, we specify task and prompt categories rather than treating arbitrary informal relations as morphisms.

\begin{definition}[Categories of Tasks and Prompts]
Let $\mathcal{T}$ be a category of typed task specifications. Its objects are task schemas, and a morphism $f:X\to Y$ is an explicitly specified, composable transformation or reduction from schema $X$ to schema $Y$. Identity morphisms are no-op transformations, and composition is sequential application. Only transformations closed under these operations are included; an informal relationship between two tasks is not automatically a morphism.

Let $\mathcal{P}$ be a concrete category of typed prompt spaces. An object $P$ is a set of prompt instances satisfying a fixed schema, and a morphism $u:P\to Q$ is a schema-preserving prompt transformation. Identities and composition are the ordinary identity functions and function composition. For the Writer-monad construction in Sec.~\ref{sec:recursive-mp}, we additionally assume that the logged spaces $P\times\Sigma$ and the maps used below are objects and morphisms of $\mathcal{P}$; a full subcategory of $\mathbf{Set}$ closed under products with the fixed edit-script set $\Sigma$ is one sufficient setting.
\end{definition}

The core structural object is a functorial meta-prompting scheme.

\begin{definition}[Meta Prompting Functor]
A \emph{Meta Prompting functor} $\mathcal{M}:\mathcal{T}\to\mathcal{P}$ consists of
\begin{itemize}[parsep=1pt, itemsep=1pt, topsep=1pt, leftmargin=*]
\item \textbf{an object assignment:} each task schema $X$ is assigned a prompt space $\mathcal{M}(X)$; and
\item \textbf{a morphism assignment:} each task transformation $f:X\to Y$ is assigned a schema-preserving prompt transformation $\mathcal{M}(f):\mathcal{M}(X)\to\mathcal{M}(Y)$.
\end{itemize}
These assignments define a functor only if, for every object $X$ and every composable pair $f:X\to Y$ and $g:Y\to Z$,
\[
\mathcal{M}(\operatorname{id}_X)=\operatorname{id}_{\mathcal{M}(X)}
\quad\text{and}\quad
\mathcal{M}(g\circ f)=\mathcal{M}(g)\circ\mathcal{M}(f).
\]
Thus, functoriality is a condition to be verified or enforced by construction; it is not guaranteed for an arbitrary prompt-generation procedure.
\end{definition}

Meta Prompting provides a systematic vocabulary for relating task decompositions to prompt decompositions. The equality above is structural equality in $\mathcal{P}$ (or equality after a declared prompt-normalization procedure). It does not imply that an LLM follows the prompt faithfully or produces a semantically correct answer.

\paragraph{Compiler view.} A task schema may contain stages such as \texttt{parse}, \texttt{compute}, and \texttt{verify}; a prompt schema contains corresponding typed sections. A functorial compiler maps a sequential task transformation to the same sequential composition of schema-preserving prompt transformations.

\begin{proposition}[Compositionality of Meta Prompting]
\label{prop:compositionality}
Let $\mathcal{M}:\mathcal{T}\to\mathcal{P}$ be a Meta Prompting functor. If $h=g\circ f$ for morphisms $f:T_1\to T_2$ and $g:T_2\to T_3$, then
\[
\mathcal{M}(h)=\mathcal{M}(g)\circ\mathcal{M}(f).
\]
\end{proposition}
\begin{proof}
Because $h=g\circ f$, the result is exactly the composition-preservation axiom in the definition of a functor. The statement concerns equality of prompt transformations in $\mathcal{P}$, not equality of downstream model behavior.
\end{proof}

A functorial mapping may be designed by a human, synthesized by a program, or proposed by an LLM and then checked against the identity and composition laws. RMP, introduced in Sec.~\ref{sec:recursive-mp}, concerns repeated edits to prompt instances and is logically separate from the existence of the functor $\mathcal{M}$.

\paragraph{Example of Meta Prompting.} Consider the task schema of solving $ax^2+bx+c=0$ for real coefficients $a,b,c$ with $a\neq0$, represented by an object $Q\in\operatorname{Ob}(\mathcal{T})$. The functor assigns a prompt space $\mathcal{M}(Q)$, and a particular template $p_Q\in\mathcal{M}(Q)$ (Fig.~\ref{fig:meta-prompt-quadratic}) can specify steps for identifying coefficients, computing the discriminant, and applying the quadratic formula. The template is applicable to any instance satisfying the declared coefficient assumptions; whether an LLM follows it correctly remains empirical.

\begin{figure}[htb]
\centering
\begin{tcolorbox}[width=0.9\textwidth, colback=blue!2!white, colframe=gray!50!blue]
\begin{minipage}{\textwidth}
\color{systemcolor}
A structured meta-prompt for solving quadratic equations of the form $ax^2 + bx + c = 0$, where $a,b,c\in\mathbb{R}$ and $a\neq0$:

\{

\hspace{4ex}``Problem'': ``Solve the quadratic equation $ax^2 + bx + c = 0$ for $x$, assuming $a,b,c\in\mathbb{R}$ and $a\neq0$.'',

\hspace{4ex}``Solution'': \{

\hspace{8ex}``Step 1'': ``Identify the coefficients $a$, $b$, and $c$, and verify that $a\neq0$.'',

\hspace{8ex}``Step 2'': ``Compute the discriminant using $\Delta = b^2 - 4ac$.'',

\hspace{8ex}``Step 3'': ``Determine the nature of the roots by checking if $\Delta > 0$, $\Delta = 0$, or $\Delta < 0$.'',

\hspace{8ex}``Step 4'': ``If $\Delta > 0$, calculate the two distinct real roots using $x_{1,2} = \frac{-b \pm \sqrt{\Delta}}{2a}$.'',

\hspace{8ex}``Step 5'': ``If $\Delta = 0$, calculate the repeated real root using $x = \frac{-b}{2a}$.'',

\hspace{8ex}``Step 6'': ``If $\Delta < 0$, calculate the complex roots using $x_{1,2} = \frac{-b \pm i\sqrt{|\Delta|}}{2a}$.'',

\hspace{8ex}``Step 7'': ``Conclude by summarizing the roots in a LaTeX-formatted box, $\boxed{\cdots}$.''

\hspace{4ex}\},

\hspace{4ex}``Final Answer'': ``State the complete real or complex root set explicitly.''
\}

----------

\end{minipage}
\end{tcolorbox}
\caption{A structured meta-prompt in JSON-like format for solving quadratic equations.}
\label{fig:meta-prompt-quadratic}
\end{figure}

\begin{figure}[htb]
\centering
\begin{tcolorbox}[width=0.9\textwidth, colback=cyan!2!white, colframe=gray!50!cyan]
\begin{minipage}{\textwidth}
\small
Example 1:

Problem: Solve the quadratic equation $3x^2 + 4x - 5 = 0$.

Solution:
\begin{itemize}[leftmargin=15pt]
\item Step 1: Identify the coefficients: $a=3$, $b=4$, and $c=-5$.
\item Step 2: Calculate the discriminant: $\Delta = b^2 - 4ac = 4^2 - 4(3)(-5) = 16 + 60 = 76$.
\item Step 3: Since $\Delta > 0$, the equation has two distinct real roots.
\item Step 4: Compute the roots using the quadratic formula: $x_{1,2} = \frac{-b \pm \sqrt{\Delta}}{2a} = \frac{-4 \pm \sqrt{76}}{6}$.
\item Step 5: Since $\sqrt{76}=2\sqrt{19}$, simplify the roots to $x_1 = \frac{-2 + \sqrt{19}}{3}$ and $x_2 = \frac{-2 - \sqrt{19}}{3}$.
\end{itemize}

Final Answer: The roots of the equation $3x^2 + 4x - 5 = 0$ are $x_1 = \frac{-2 + \sqrt{19}}{3}$ and $x_2 = \frac{-2 - \sqrt{19}}{3}$.

----------

Example 2: \ldots
\end{minipage}
\end{tcolorbox}
\caption{An example of a few-shot prompt for solving quadratic equations with detailed steps.}
\label{fig:fewshot-prompt-quadratic}
\end{figure}

\paragraph{Characteristics of Meta Prompting.} Meta Prompting prioritizes form and structure over content by using a syntactic template for the expected response. Placeholders describe the roles of problem statements, intermediate steps, checks, and conclusions without providing worked task instances. The type-theoretic analogy is operational only when these fields are parsed or validated; otherwise, it remains an informal specification expressed in natural language.

\subsection{Distinctions between Meta Prompting and Few-Shot Prompting}

Meta Prompting differs from few-shot prompting in the information placed in context. Few-shot prompting supplies concrete, content-rich ``(problem, solution)'' pairs that support in-context analogy. Meta Prompting instead supplies an example-free structural template that specifies the desired decomposition and output format. The distinction is about prompt content and organization, not a claim that the template reveals or controls the model's internal reasoning process.

Beyond few-shot methods, Meta Prompting also distinguishes itself from other programmatic or structured prompting approaches, such as those using XML tags or frameworks like DSPy~\citep{khattab2023dspy}. While these methods also impose structure, they often function as programming layers that compile into traditional few-shot or zero-shot prompts. Meta Prompting, as formalized here, is a more fundamental concept focused on the direct, example-agnostic mapping between a task's abstract structure and a prompt's syntactic structure. For further illustration of these differences, please refer to Figures~\ref{fig:meta-prompt-math-json},~\ref{fig:meta-prompt-math-markdown}, and~\ref{fig:fewshot-prompt-math}.

\subsection{Meta Prompting for Complex Reasoning}

Structured prompts can make interfaces to symbolic systems and code environments easier to parse and validate. In mathematics and logic, explicit fields for assumptions, intermediate claims, code, outputs, and final answers can support tool orchestration and error checking. These benefits are interface-level hypotheses rather than formal consequences of the categorical model; their effect on accuracy must be measured empirically. Figure~\ref{fig:mp-cr} in Appendix~\ref{sec:examples} gives an illustrative template.

\subsection{Advantages and Scope of Meta Prompting}

Meta Prompting can offer two practical advantages over long few-shot prompts, subject to the model and task.

\noindent\textbf{Token efficiency.} Replacing worked examples with a compact schema can reduce prompt length, especially when the same schema is reused across many instances. This is not automatic: an elaborate schema can be longer than a small few-shot prompt, and generated reasoning may dominate the total token budget.

\noindent\textbf{Example-free evaluation.} Because MP contains no worked exemplars, it reduces sensitivity to exemplar selection and avoids direct leakage from those exemplars~\citep{brown2020language, liu2020multi, reynolds2021prompt}. It does not make evaluation unbiased: results can still depend on the wording of the template, the model's training data, decoding settings, and the evaluator.

Accordingly, the relevant empirical questions are whether a given schema improves accuracy, robustness, or token use under a controlled protocol, and how sensitive those outcomes are to alternative schemas.

\begin{algorithm}[H]
\caption{Recursive Meta Prompting (RMP)}
\label{alg:rmp}
\begin{algorithmic}[1]
\Require Task $t$, meta-meta-prompt $p_{\mathrm{meta}}$, proposer $\mathcal{L}_{\mathrm{prop}}$, executor $\mathcal{L}_{\mathrm{exec}}$, maximum iterations $K$
\State $p \gets \operatorname{InitialPrompt}(t)$
\For{$i=1$ \textbf{to} $K$}
  \State $\widetilde{s} \gets \mathcal{L}_{\mathrm{prop}}(p_{\mathrm{meta}},t,p)$ \Comment{Propose an edit}
  \State $s \gets \operatorname{ValidateEdit}(p,\widetilde{s})$
  \If{$s=\bot$}
    \State \textbf{break} \Comment{Reject an invalid edit}
  \EndIf
  \State $p' \gets \operatorname{Apply}(p,s)$
  \State $c \gets \operatorname{IsConverged}(p',p)$
  \State $p \gets p'$ \Comment{Retain the final candidate even when stopping}
  \If{$c$}
    \State \textbf{break}
  \EndIf
\EndFor
\State \Return $\mathcal{L}_{\mathrm{exec}}(p,t)$
\end{algorithmic}
\end{algorithm}

\section{Recursive Meta Prompting: Self-Refinement and Automation}
\label{sec:recursive-mp}

Meta-prompts can also be used to generate and refine other prompts. We call this process \textbf{Recursive Meta Prompting (RMP)}. RMP separates a proposer, which emits prompt edits, from an executor, which solves tasks with the resulting prompt. This resembles metaprogramming, where programs are represented as data that other programs can inspect and transform. We use the term ``refinement'' operationally; no monotonic improvement guarantee is assumed.

\begin{figure}[!ht]
\centering
\resizebox{0.72\textwidth}{!}{%
\begin{tikzpicture}[
node distance=1cm and 1.5cm,
auto,
thick,
main node/.style={circle, draw, fill=blue!10, font=\sffamily\small\bfseries, text width=1.5cm, align=center, minimum size=2cm},
llm node/.style={rectangle, draw, fill=blue!35, font=\sffamily\small\bfseries, text width=1.5cm, align=center, minimum size=2cm},
meta node/.style={circle, draw, fill=purple!70, font=\sffamily\small\bfseries, text width=1.5cm, align=center, minimum size=2cm},
line/.style={-Stealth, thick},
]
\node[main node] (task) {Task};
\node[meta node] (metaMeta) [above right=of task] {Meta\\Meta\\Prompt};
\node[llm node] (proposer) [below right=of metaMeta] {Meta\\Prompt\\Proposer};
\node[meta node] (generated) [above right=of proposer] {New\\Meta\\Prompt};
\node[llm node] (executor) [below right=of generated] {Meta\\Prompt\\Executor};
\node[main node] (solved) [below right=of executor] {Draft\\Answer};
\draw[line] (task) -- (proposer);
\draw[line] (metaMeta) -- (proposer);
\draw[line] (proposer) -- (generated);
\draw[line] (generated) -- (executor);
\draw[line] (executor) -- (solved);
\draw[line] (task) to[bend right=45] (executor);
\end{tikzpicture}
}
\caption{The workflow of Recursive Meta Prompting. A \textit{meta-meta-prompt} guides a proposer LLM to generate a task-specific meta-prompt, which an executor LLM then uses to produce a candidate solution to the original task.}
\label{fig:meta-meta-prompt}
\end{figure}

\subsection{A Monadic Framework for Prompt Refinement}

We use a Writer monad to model the accumulation of prompt-edit scripts. This construction records how edits compose; a separate action applies the accumulated script to a prompt. Let $\mathcal{P}$ be the concrete category of typed prompt spaces and schema-preserving functions introduced above. Fix an edit-script set $\Sigma$ and assume that the logged spaces and maps defined below lie in $\mathcal{P}$; this holds, for example, in a suitable full subcategory of $\mathbf{Set}$ closed under products with $\Sigma$. Let $(\Sigma,\cdot,e)$ be a monoid of valid edit scripts, where $s_1\cdot s_2$ means ``apply $s_1$ and then $s_2$,'' and $e$ is the empty edit.

Define the Writer endofunctor $\mathsf{W}:\mathcal{P}\to\mathcal{P}$ by
\[
\mathsf{W}(X)=X\times\Sigma,
\qquad
\mathsf{W}(f)(p,s)=(f(p),s)
\]
for every schema-preserving function $f:X\to Y$. Its unit and multiplication have components
\begin{align*}
\eta_X(p)&=(p,e),\\
\mu_X\bigl((p,s_1),s_2\bigr)&=(p,s_1\cdot s_2).
\end{align*}
The monoid identity and associativity laws imply the monad unit and associativity laws, respectively; naturality follows from the componentwise definitions.

The Writer monad stores edits but does not itself modify the prompt component $p$. To model execution of edits on a fixed prompt space $X$, assume a schema-preserving right action
\[
\rho_X:X\times\Sigma\to X
\]
satisfying
\[
\rho_X(p,e)=p,
\qquad
\rho_X\bigl(\rho_X(p,s_1),s_2\bigr)=\rho_X(p,s_1\cdot s_2).
\]
Equivalently, $\rho_X$ is an algebra for the Writer monad. These equations state that applying an accumulated script agrees with sequential application. They do not state that two edits commute, that refinement converges, or that the edited prompt is more accurate.

\begin{proposition}[Associativity of Edit Accumulation]
\label{prop:stability}
For any prompt $p\in X$ and edit scripts $s_1,s_2,s_3\in\Sigma$, the two ways of flattening an element of $\mathsf{W}^3(X)$ produce the same accumulated script:
\[
(p,(s_1\cdot s_2)\cdot s_3)=(p,s_1\cdot(s_2\cdot s_3)).
\]
Consequently, the result is independent of parenthesization, though generally not of the order of the edits.
\end{proposition}

\begin{proof}
For $(((p,s_1),s_2),s_3)\in\mathsf{W}^3(X)$, the path $\mu_X\circ\mathsf{W}(\mu_X)$ yields $(p,(s_1\cdot s_2)\cdot s_3)$, whereas the path $\mu_X\circ\mu_{\mathsf{W}(X)}$ yields $(p,s_1\cdot(s_2\cdot s_3))$. These are equal by associativity in the monoid $\Sigma$.
\end{proof}

\paragraph{A concrete walkthrough of RMP.} Let $p_0$ be an initial prompt for task $t$. At iteration $i$, the proposer emits an edit script $s_i$, and the prompt is updated by
\[
p_{i+1}=\rho_X(p_i,s_i).
\]
After $K$ updates,
\[
p_K=\rho_X\bigl(p_0,s_0\cdot s_1\cdot\cdots\cdot s_{K-1}\bigr),
\]
with any parenthesization of the product giving the same result. A convergence test, if used, is an external stopping rule and is not supplied by the monad laws.

\paragraph{Computational costs.} RMP adds proposer calls during prompt construction. When a refined prompt is reused across many downstream instances, this offline cost can be amortized; when prompts are refined per instance, it cannot. The executor stage can still use a single compact prompt per downstream instance.

Operationally, the proposer--executor loop can be written as
\begin{align*}
s_i &= R(p_{\mathrm{meta}},t,p_i), & p_{i+1}&=\rho_X(p_i,s_i), & \widehat{y}&=E(t,p_K),
\end{align*}
where $R$ is the proposer and $E$ is the executor. The algebraic model governs edit composition, not the semantic quality of $\widehat{y}$.

\subsection{Case Study: Automatic Prompt Derivation}

To make the RMP process concrete, consider deriving a prompt for a new problem domain. Algorithm~\ref{alg:rmp} describes the proposer--executor loop used to generate candidate meta-prompts. For instance, an LLM equipped with a high-level ``meta-meta-prompt'' (Fig.~\ref{fig:rmp-meta-meta-example}) can take a current prompt as input and propose a schema-preserving revision.

Figure~\ref{fig:rmp-meta-meta-example} gives an example of a meta-meta-prompt. Given a document or current prompt, it instructs the proposer to analyze the task and emit a new structured prompt. Repeating this step implements the operational loop in Algorithm~\ref{alg:rmp}; stabilization and downstream accuracy remain separate empirical questions.

\begin{figure}[ht!]
\centering
\begin{tcolorbox}[width=\textwidth, colback=blue!2!white, colframe=gray!50!blue]
\textbf{Task:} \textit{Meta Prompting for In-Context Prompt Design}
\begin{enumerate}[leftmargin=15pt]
\item \textbf{Document Analysis:}
\begin{itemize}[leftmargin=15pt]
\item \textbf{Input:} [Complex document (e.g., a research paper or this prompt itself)]
\item \textbf{Action:} Analyze and extract key concepts, methodologies, challenges, and objectives.
\end{itemize}
\item \textbf{Task Interpretation:}
\begin{itemize}[leftmargin=15pt]
\item \textbf{Action:} Synthesize the extracted information to define the core problem or task.
\item \textbf{Considerations:} Identify constraints, goals, or requirements.
\end{itemize}
\item \textbf{Prompt Design:}
\begin{itemize}[leftmargin=15pt]
\item \textbf{Objective:} Develop a structured prompt for problem-solving, including clear instructions, a step-by-step approach, and relevant background information.
\end{itemize}
\item \textbf{Optional – Direct Solution Proposal:}
\begin{itemize}[leftmargin=15pt]
\item \textbf{Objective:} Propose initial steps or a complete solution strategy, ensuring feasibility and practicality.
\end{itemize}
\item \textbf{Output Prompt:} [Generate the output prompt using the same LaTeX format as this template.]
\end{enumerate}
\textit{Note: The output should be a coherent, actionable prompt or solution strategy tailored to the specifics of the input document.}
\end{tcolorbox}
\caption{An example of a meta-meta-prompt for In-Context Prompt Design (MP-ICPD). This prompt instructs an LLM on how to analyze a document and generate a new, structured meta-prompt to solve the task described within it.}
\label{fig:rmp-meta-meta-example}
\end{figure}

By automating prompt proposal and revision, RMP can reduce manual prompt-engineering effort. Any gain in adaptability or task performance remains an empirical property of the proposer, executor, stopping rule, and evaluator.

\section{Experiments}
\label{sec:experiments}

In this section, we evaluate the performance of our proposed Meta Prompting (MP) framework on several mathematical benchmarks and problem-solving tasks. Our experiments are designed to assess both accuracy and efficiency.

\subsection{Solving MATH and GSM8K Problems}

\noindent\textbf{Experimental setup.} We evaluate on two standard test sets. MATH~\citep{hendrycks2021measuring} contains 5,000 competition-level problems, and GSM8K~\citep{cobbe2021training} contains 1,319 grade-school mathematics problems. We perform inference with vLLM using Qwen-14B and Qwen-72B \emph{base} models.

Each benchmark uses a fixed, task-specific meta-prompt generated by RMP from the same task-agnostic meta-meta-prompt (Sec.~\ref{sec:recursive-mp}; artifacts in Appendix~\ref{sec:examples} and the Supplement). For MATH, we use the markdown template in Fig.~\ref{fig:meta-prompt-math-markdown}; for GSM8K, we use the JSON-like template in Fig.~\ref{fig:meta-prompt-math-json}.

We score outputs with a rule-based evaluator that combines formatting normalization with SymPy~\citep{sympy} equivalence checks. With one fixed decode per item, pass@1 is the empirical fraction of test items scored correct. For descriptive item-level uncertainty, we report the two-sided 95\% Wald interval
\[
\widehat{p}\;\pm\;1.96\sqrt{\frac{\widehat{p}(1-\widehat{p})}{n}}.
\]
This interval treats benchmark items as independent Bernoulli trials. It does not capture decoding randomness, prompt-selection uncertainty, evaluator error, or uncertainty over the choice of model.

\noindent\textbf{Experimental results.} With one fixed structure-only template per benchmark, the Qwen-72B base model obtains \textbf{46.3\%} pass@1 on MATH (95\% interval: \textbf{[44.9, 47.7]}; $n{=}5000$) and \textbf{83.5\%} on GSM8K (95\% interval: \textbf{[81.5, 85.5]}; $n{=}1319$). We present these as proof-of-concept results and discuss template and model sensitivity in Sec.~\ref{sec:limitations}. The cross-model entries in Tables~\ref{tab:math-acc} and~\ref{tab:gsm8k-acc} are contextual comparisons rather than a controlled leaderboard: training data, prompts, decoding, and evaluators can differ across reports.

\begin{table*}[htb!]
\caption{Reported pass@1 accuracy on MATH without external tool use. Values outside our Qwen experiments are reproduced from the cited sources and may use different prompts, decoding settings, and evaluators; they should therefore be read as context rather than as a controlled comparison.}
\label{tab:math-acc}
\begin{center}
\begin{small}
\begin{tabular}{lcccr}
\toprule
Model  & Fine-Tuning Data & Tool Usage & Eval Method  &  MATH (\%) \\
\midrule
\multicolumn{5}{l}{\textbf{Proprietary Models}} \\
\midrule
\quad Claude-2~\citep{anthropic2023claude} & - & No & CoT & 32.5\\
\quad Minerva-540B~\citep{lewkowycz2022solving} & arXiv+Web & No & CoT  & 33.6 \\
\quad PaLM-2~\citep{anil2023palm} & - & No & CoT & 34.3\\
\quad GPT-4 {\tiny(\texttt{gpt-4-0314})}~\citep{OpenAI2023GPT4TR} & - & No & CoT  & 42.5 \\
\midrule
\multicolumn{5}{l}{\textbf{Open-weight Models}} \\
\midrule
\quad Qwen-14B (base)  & - & No  & CoT  & \underline{24.8} \\
\quad Qwen-14B (base)  & - & No  & \textbf{MP} & \textbf{28.9} \\
\quad Qwen-72B (base)  & - & No  & CoT  & \underline{35.2} \\
\quad Qwen-72B-MetaMathQA  & MetaMathQA & No  & CoT  & 41.7 \\
\quad Qwen-72B (base)  & - & No  & \textbf{MP} & \textbf{46.3} \\
\bottomrule
\end{tabular}
\end{small}
\end{center}
\end{table*}

\begin{table*}[htb!]
\caption{Reported pass@1 accuracy on GSM8K without external tool use. Values from other systems are reproduced from the cited sources and may use different prompts, decoding settings, and evaluators. The paired Qwen base-model rows provide the most direct within-model comparison.}
\label{tab:gsm8k-acc}
\begin{center}
\begin{small}
\begin{tabular}{lcccr}
\toprule
Model  & Fine-Tuning Data & Tool Usage & Eval Method  &  GSM8K (\%) \\
\midrule
\quad Qwen-14B (base)~\citep{bai2023qwen}  & - & No  & CoT  & \underline{61.3} \\
\quad Qwen-14B (base)  & - & No  & \textbf{MP} & \textbf{64.8} \\
\quad WizardMath-70B~\citep{luo2023wizardmath} & WizardMath & No & CoT & 81.6 \\
\quad MetaMath-70B~\citep{yu2023metamath} & MetaMathQA & No & CoT & 82.3 \\
\quad Qwen-72B (base)  & - & No  & CoT  & \underline{78.9} \\
\quad Qwen-72B (base)  & - & No  & \textbf{MP} & \textbf{83.5} \\
\bottomrule
\end{tabular}
\end{small}
\end{center}
\end{table*}

\subsection{Solving the Game of 24 Tasks}

\noindent\textbf{Task and validation.} In the Game of 24~\citep{yao2023tree}, a valid expression must use each of the four input numbers exactly once, use only the binary operations $+$, $-$, $\times$, and $\div$ with arbitrary parenthesization, avoid division by zero, and evaluate exactly to $24$. The corrected evaluation protocol uses rational arithmetic. A puzzle is counted as solved only when an expression parser verifies all of these conditions; checking merely that a solution field is non-empty is insufficient because a non-empty failure message would be a false positive.

\noindent\textbf{Protocol.} The MP-CR (Meta Prompting for Complex Reasoning) system uses the prompt in Fig.~\ref{fig:mp-cr-xml-0.2} to synthesize one Python solver and then executes that solver over a batch of $N{=}1362$ puzzles. This differs materially from IO, CoT, and ToT~\citep{yao2023tree}, which use model generations during the solution of each puzzle. Consequently, Table~\ref{tab:gameof24-comparison} is a resource-accounting comparison across different protocols, not a controlled comparison of prompting strategies.

\noindent\textbf{Reported results.} The original workflow reports a 100\% success rate on 1,362 puzzles. The corrected exact-arithmetic solver and verifier in Appendix~\ref{sec:more-game-of-24} define how that claim should be recomputed, but the aggregate cannot be independently confirmed without the input CSV and per-instance output. The reported average processing time is 0.08 seconds per puzzle in the OpenAI Assistant workflow. A single batch response used approximately 8,000 generated tokens and 1,000 prompt tokens, which amortizes to
\[
\frac{8000}{1362}\approx 5.87
\quad\text{generated tokens and}\quad
\frac{1000}{1362}\approx 0.73
\quad\text{prompt tokens per puzzle}.
\]
These figures characterize a batch program-synthesis workflow; they do not establish that MP is uniformly more accurate or efficient under matched per-instance inference protocols.

\begin{table*}[ht!]
\centering
\small
\vspace{-1ex}
\caption{Reported resource use and success rates for Game of 24. IO, CoT, and ToT solve puzzles through per-instance model generation, whereas MP-CR synthesizes one program and executes it over the full batch. The protocols are therefore not directly comparable. An API call denotes one complete query--response. MP-CR tokens, calls, and cost are amortized over $N{=}1362$ puzzles.}
\label{tab:gameof24-comparison}
\resizebox{\textwidth}{!}{%
\begin{tabular}{@{}lcccc@{}}
\toprule
Method & LLM Calls per Puzzle & Gen./Prompt Tokens per Puzzle & Cost per Puzzle (USD) & Success Rate \\
\midrule
IO (best of 100) & 100 & 1.8k / 1.0k & 0.13 & 33\% \\
CoT (best of 100) & 100 & 6.7k / 2.2k & 0.47 & 49\% \\
ToT~\citep{yao2023tree} & 61.72 & 5.5k / 1.4k & 0.74 & 74\% \\
\textbf{MP-CR (batch program)} & $1/1362$ & $\approx 5.87 / 0.73$ & $\approx 0.0003$ & \textbf{100\% (reported)} \\
\bottomrule
\end{tabular}%
}
\end{table*}

\section{Related Work}

\paragraph{Reasoning with AI Systems.}
Efforts to enhance AI reasoning capabilities have largely focused on equipping neural networks with mechanisms to generate intermediate reasoning steps, a strategy that has yielded improvements across diverse domains~\citep{zaidan2007using, yao2021refining, hase2021can, yang2022seqzero, wu2022ai, zhou2022least}. Although these approaches have advanced the state of the art, they predominantly emphasize content-driven reasoning. In parallel, substantial research has investigated the use of symbolic systems, such as code environments and knowledge graphs, to further augment reasoning~\citep{mihaylov2018knowledgeable, bauer2018commonsense, kundu2018exploiting, wang2019improving, lin2019kagnet, ding2019cognitive, feng2020scalable, wang2022multi, chen2022program, lyu2023faithful, gao2023pal, gou2023tora, Jiang2022DraftSA, yang2023leandojo}. In contrast, our work on meta prompting shifts the focus from content-centric methods to a structural and formal treatment of reasoning processes.

\paragraph{Chain-of-Thought Prompting.}
The introduction of Chain-of-Thought (CoT) prompting by \citet{wei2022chain} marked a significant milestone by emphasizing the articulation of intermediate reasoning steps. This foundational idea has been extended in numerous ways. Methodologies like Self-Consistency~\citep{wang2022self} and Complex CoT~\citep{fu2022complexity} focus on generating multiple reasoning chains and selecting the best one, often through voting. Decomposition strategies, such as Least-to-Most~\citep{zhou2022least} and Decomposed Prompting~\citep{khot2022decomposed}, focus on breaking complex problems into simpler, solvable sub-tasks. More recent work has explored multi-agent debates~\citep{du2023improving}, diverse reasoning paths with verifiers~\citep{li2023making}, and progressive, iterative refinement~\citep{zheng2023progressive}. A significant parallel thread enables LLMs to self-criticize and self-correct their reasoning paths, accompanied in some settings by theoretical analyses of improvement~\citep{tyen2023llms, li2024confidence, wang2024theoretical}. These methods primarily target the semantic content of a reasoning trace and may rely on few-shot examples or external feedback. Meta Prompting instead specifies the external structure and interfaces of the prompt; it does not by itself control the model's latent reasoning process.

\paragraph{Structured and Graph-Based Reasoning Frameworks.} Recognizing the limitations of linear reasoning chains, recent work has explored more complex reasoning topologies. The Tree-of-Thought (ToT) framework~\citep{yao2023tree, long2023large} was a significant step, allowing an LLM to explore multiple reasoning paths in a tree structure and use self-evaluation to prune branches. This idea has been generalized by Cumulative Reasoning (CR)~\citep{zhang2023cumulative}, Graph-of-Thoughts (GoT)~\citep{besta2024graph}, and Diagram-of-Thought (DoT)~\citep{zhang2024diagram}, which represent thoughts and dependencies with graph-like structures, as well as Forest-of-Thoughts (FoT)~\citep{bi2024forest}, which explores diverse high-level plans concurrently. These frameworks provide powerful high-level strategies for exploring a problem space. Our work on Meta Prompting is orthogonal and potentially complementary. While ToT, GoT, and FoT define a macro-level topology for search (for example, a tree or graph), Meta Prompting specifies schemas for the prompts associated with nodes and transformations. A concrete integration would still need to define and verify the relevant categories, morphisms, and compiler.

\section{Conclusion}

We introduced Meta Prompting as an example-free approach to specifying the external structure of a task solution. Under explicitly defined categories of typed task schemas and typed prompt spaces, a Meta Prompting functor makes identity and composition preservation precise design conditions. For Recursive Meta Prompting, the Writer monad records ordered edit histories, while a separate monoid action applies those edits to prompts. The monad laws guarantee identity and independence from parenthesization of edit accumulation; they do not guarantee convergence, semantic correctness, commutativity of edits, or monotonic improvement.

Our proof-of-concept experiments report competitive results for Qwen base models on MATH and GSM8K under one fixed structure-only prompt per benchmark. The Game of 24 study is a separate batch program-synthesis workflow and should not be interpreted as a matched comparison with per-instance IO, CoT, or ToT inference. More broadly, the formalism is best viewed as a specification language for modular prompt transformations. Establishing practical value requires controlled ablations over templates and recursion depth, explicit validation of generated edits, reproducible evaluators, and matched comparisons across models and inference protocols.

\vspace{5ex}
\bibliography{reference}
\bibliographystyle{plainnat}

\clearpage
\appendix

\renewcommand{\appendixpagename}{\centering \LARGE Appendix}
\appendixpage

\startcontents[section]
\printcontents[section]{l}{1}{\setcounter{tocdepth}{2}}
\clearpage

\section{Theoretical Foundations}
\label{sec:appendix-theory}

\subsection{Type Theory}
\label{sec:intro-type-theory}

Type theory is a family of formal systems in which expressions are classified by types. Examples include the simply typed $\lambda$-calculus and dependent type theories such as Martin-Löf type theory. Different type theories make different choices about logic, computation, equality, inductive definitions, and classical principles; there is no single collection of properties shared by every system called ``type theory.''

A typing judgment is commonly written $\Gamma\vdash t:A$, meaning that the term $t$ has type $A$ under context $\Gamma$. In the simply typed $\lambda$-calculus, if $\Gamma,x:A\vdash t:B$, then the lambda abstraction satisfies
\[
\Gamma\vdash (\lambda x:A.\,t):A\to B.
\]
Many computational type theories also equip terms with reduction rules, so syntactically different terms can be definitionally equal after normalization. Other type-theoretic settings may use a different notion of equality or may be studied primarily through semantics.

The connection to Meta Prompting is an analogy unless the prompt is coupled to a formal grammar or validator. A schema may assign type-like roles to fields such as \texttt{ProblemStatement}, \texttt{ReasoningSteps}, and \texttt{FinalAnswer}; a parser or constrained decoder can then check whether an output inhabits the declared schema. Natural-language instructions alone do not provide the guarantees of a formal type system.

\subsection{Proof of Compositionality (Proposition~\ref{prop:compositionality})}
\begin{proof}
Let $f:T_1\to T_2$ and $g:T_2\to T_3$ be composable morphisms in $\mathcal{T}$, and let $h=g\circ f$. Since $\mathcal{M}:\mathcal{T}\to\mathcal{P}$ is assumed to be a functor,
\[
\mathcal{M}(h)
=\mathcal{M}(g\circ f)
=\mathcal{M}(g)\circ\mathcal{M}(f).
\]
The first equality substitutes the definition of $h$, and the second is the functor composition law. This is a structural equality of morphisms in $\mathcal{P}$. It does not imply that an executor LLM follows either prompt transformation faithfully or that the resulting answers are semantically equivalent.
\end{proof}

\subsection{The Writer Monad for Edit Histories}
\label{sec:appendix-monad}

Section~\ref{sec:recursive-mp} models the edit-history component of RMP with the Writer monad $\mathsf{W}$. Let $(\Sigma,\cdot,e)$ be a monoid, and let $\mathcal{P}$ be closed under products with $\Sigma$. Recall
\[
\mathsf{W}(X)=X\times\Sigma,
\qquad
\mathsf{W}(f)(p,s)=(f(p),s),
\]
with
\[
\eta_X(p)=(p,e),
\qquad
\mu_X((p,s_1),s_2)=(p,s_1\cdot s_2).
\]
The monad laws are
\begin{align}
\mu_X\circ\mathsf{W}(\eta_X)&=\operatorname{id}_{\mathsf{W}(X)},
&
\mu_X\circ\eta_{\mathsf{W}(X)}&=\operatorname{id}_{\mathsf{W}(X)},
\notag\\
\mu_X\circ\mathsf{W}(\mu_X)&=\mu_X\circ\mu_{\mathsf{W}(X)}.
\label{eq:monad-laws}
\end{align}

\paragraph{Unit laws.}
For $(p,s)\in\mathsf{W}(X)$,
\begin{align*}
(\mu_X\circ\mathsf{W}(\eta_X))(p,s)
&=\mu_X((p,e),s)
=(p,e\cdot s)
=(p,s),\\
(\mu_X\circ\eta_{\mathsf{W}(X)})(p,s)
&=\mu_X((p,s),e)
=(p,s\cdot e)
=(p,s).
\end{align*}
These equalities use the left and right identity laws of the monoid.

\paragraph{Associativity.}
For $(((p,s_1),s_2),s_3)\in\mathsf{W}^3(X)$,
\begin{align*}
(\mu_X\circ\mathsf{W}(\mu_X))(((p,s_1),s_2),s_3)
&=(p,(s_1\cdot s_2)\cdot s_3),\\
(\mu_X\circ\mu_{\mathsf{W}(X)})(((p,s_1),s_2),s_3)
&=(p,s_1\cdot(s_2\cdot s_3)).
\end{align*}
The two results are equal by associativity of $\cdot$.

\paragraph{Naturality.}
For a morphism $f:X\to Y$,
\[
\mathsf{W}(f)(\eta_X(p))=(f(p),e)=\eta_Y(f(p)),
\]
so $\eta$ is natural. Likewise, both paths in the naturality square for $\mu$ send $((p,s_1),s_2)$ to $(f(p),s_1\cdot s_2)$, so $\mu$ is natural.

\subsection{Verification of Associativity of Edit Accumulation}

Proposition~\ref{prop:stability} is exactly the componentwise associativity calculation above. It establishes independence from \emph{parenthesization}. It does not establish independence from edit order: unless $\Sigma$ is commutative,
\[
s_1\cdot s_2\neq s_2\cdot s_1
\]
in general. Nor does associativity imply termination, convergence to a fixed point, or improvement in executor accuracy.

To apply the recorded history to a prompt space $X$, the action $\rho_X:\mathsf{W}(X)\to X$ must satisfy the Writer-algebra laws
\[
\rho_X\circ\eta_X=\operatorname{id}_X,
\qquad
\rho_X\circ\mathsf{W}(\rho_X)=\rho_X\circ\mu_X.
\]
Componentwise, these are precisely
\[
\rho_X(p,e)=p,
\qquad
\rho_X(\rho_X(p,s_1),s_2)=\rho_X(p,s_1\cdot s_2).
\]

\subsection{Assumptions and Scope of the RMP Model}
\label{sec:assumptions-rmp}
The formalization uses the following assumptions:
\begin{itemize}[leftmargin=15pt]
    \item \textbf{Concrete prompt spaces.} Each object $X$ is a set of prompt instances satisfying a declared schema, and morphisms are total schema-preserving functions.
    \item \textbf{Typed edit monoid.} Valid edit scripts form a monoid $(\Sigma,\cdot,e)$ under ordered composition. The operation need not be commutative.
    \item \textbf{Product closure.} For each prompt space $X$, the logged space $X\times\Sigma$ is also an object of $\mathcal{P}$.
    \item \textbf{Well-defined action.} Script application is represented by a total action $\rho_X:X\times\Sigma\to X$ satisfying the action laws above. Invalid or partial edits must be rejected, repaired, or represented explicitly outside this model.
    \item \textbf{Structural equality.} Equalities concern prompt schemas and transformations, possibly after a declared normalization procedure; they do not identify prompts merely because an LLM happens to respond similarly to them.
    \item \textbf{External optimization criteria.} Stopping rules, convergence, semantic correctness, and measured performance are properties of the operational proposer--executor system and evaluator, not consequences of the monad laws.
\end{itemize}

\section{Additional Prompt Examples}
\label{sec:examples}

\begin{figure}[htbp]
\centering
\begin{tcolorbox}[width=0.95\textwidth, colback=gray!2!white, colframe=gray!50!blue]
\begin{minipage}{\textwidth}
\color{systemcolor}
You are ChatGPT, a state-of-the-art language model with specialized expertise in mathematics. Your strengths include tackling complex mathematical challenges using intricate reasoning and delivering solutions via methodical problem-solving. Throughout this interaction, you will encounter a variety of mathematical problems—from basic arithmetic to advanced calculus and beyond.

Your primary objective is to:
\begin{enumerate}
\item Clearly interpret and understand the problem statement.
\item Decompose the problem into manageable components, if necessary.
\item Apply appropriate mathematical principles and techniques to solve each component.
\item Synthesize the component solutions into a comprehensive answer.
\item Provide a clear, step-by-step explanation of your methodology, ensuring that your reasoning is rigorous, precise, and easily understandable.
\end{enumerate}

Your demonstrated proficiency in mathematics is expected to guide users through the problem-solving process, offering insights, strategies, and explanations that illuminate the path to the solution.
\end{minipage}
\end{tcolorbox}
\caption{An illustrative example of a generic system Meta Prompt for solving a wide range of reasoning tasks. This prompt serves as a template suitable for most tasks.}
\label{fig:system-meta-prompt}
\end{figure}

\begin{figure}[htbp]
\centering
\begin{tcolorbox}[width=0.95\textwidth, colback=blue!2!white, colframe=gray!50!blue]
\begin{minipage}{0.95\textwidth}
\centering
\tiny
\begin{lstlisting}[language=markdown,breaklines=true,columns=fullflexible,keepspaces=true]
## Problem: [problem]

Solution: Let's think step by step. [initial interpretation of the problem]

### Preliminary Content

- **Prelim 1**: [preliminary content 1]
- **Prelim 2**: [preliminary content 2]
- [...]

### Hints
- **Hint 1**: [useful hint 1]
- **Hint 2**: [useful hint 2]
- [...]

### Intermediate Steps: Question-Answer, Sketch-Code, Output, and Answer Pairs

Let's think step by step.

#### Question 1: [the first sub-question]
- **Answer Sketch**: [sketch of the answer for question 1]

##### Code for Question 1
[when applicable, execute code to check computable parts and refine your answer sketch for question 1]

#### Answer for Question 1
- **Answer**: [final answer for question 1, based on code interpreter results if available]

#### Question 2: [the second sub-question]
- **Answer Sketch**: [sketch of the answer for question 2]

##### Code for Question 2
[when applicable, execute code to check computable parts and refine your answer sketch for question 2]

#### Answer for Question 2
- **Answer**: [final answer for question 2, based on code interpreter results if available]

### [Additional Questions as Needed]

### Final Solution

Recall the original problem: <MathP> [original problem] </MathP>.

Let's think step by step.

#### Solution Sketch
[provide an overall sketch for the final solution]

#### Code for Final Solution
[when applicable, execute code to check computable parts of the solution]

#### Final Answer
[present the final answer in a LaTeX-formatted box, e.g., $\boxed{63\pi}$]
Final Answer: the answer is $\boxed{\cdots}$.

\end{lstlisting}
\end{minipage}
\end{tcolorbox}
\caption{An illustration of Meta Prompting for Complex Reasoning.}
\label{fig:mp-cr}
\end{figure}

\noindent\textbf{Key Elements of Meta Prompting for Complex Reasoning: }
\begin{enumerate}[topsep=0pt, itemsep=0.5pt, leftmargin=15pt]
\item \textbf{Complex Problem Decomposition}: Break down intricate problems into smaller, manageable sub-problems to enable systematic problem solving.
\item \textbf{Detailed Preliminary Content}: Supply essential background information and foundational concepts to set the stage for problem resolution.
\item \textbf{Step-by-Step Problem Solving}:
\begin{itemize}[topsep=0pt, itemsep=0pt, leftmargin=15pt]
\item Formulate targeted intermediate questions.
\item Develop answer sketches and use code to check computable claims when applicable.
\item Present comprehensive, step-by-step answers leading to the final solution.
\end{itemize}
\item \textbf{Final Solution Presentation}:
\begin{itemize}[topsep=0pt, itemsep=0pt, leftmargin=15pt]
\item Synthesize intermediate findings into a complete solution.
\item Check computable parts of the final solution through code execution when applicable.
\item Present the final answer in a clear and formatted manner (e.g., using $\boxed{\,}$).
\end{itemize}
\end{enumerate}

\begin{figure}[htbp]
\centering
\begin{tcolorbox}[width=0.95\textwidth, colback=blue!2!white, colframe=gray!50!blue]
\begin{minipage}{\textwidth}
\textbf{Task:} \textit{Prompt Simplification}
\begin{enumerate}[leftmargin=15pt]
\item \textbf{Original Prompt:} [input prompt]
\item \textbf{Goal:} Transform the original prompt into a concise version while preserving its core objectives.
\item \textbf{Transformation Instructions:}
\begin{enumerate}[leftmargin=15pt]
\item Retain the primary purpose and objectives.
\item Distill the prompt to include only the key instructions and essential information.
\item Eliminate extraneous details.
\item Use clear, direct language, and structure the prompt with bullet points or numbered steps for clarity.
\end{enumerate}
\item \textbf{Outcome:} The revised prompt should be succinct yet sufficiently detailed to guide effective task completion.
\end{enumerate}
\end{minipage}
\end{tcolorbox}
\caption{Illustration of Meta Prompting for designing concise prompts.}
\label{fig:special-meta-prompting-concise}
\end{figure}

\clearpage
\section{Additional Experimental Details}
\label{sec:more-experiments}

\subsection{Solving Game of 24 Tasks}
\label{sec:more-game-of-24}

This experiment is a batch program-synthesis case study. The LLM proposes a solver once, after which deterministic code searches each puzzle. Correctness should be checked independently of the solver's success flag: the verifier parses each expression, confirms that the four input numbers are used with the correct multiplicities, rejects unsupported operations and division by zero, and evaluates the expression with exact rational arithmetic. The corrected reference implementation appears in Listing~\ref{lst:mp-cr-gameof24}. The original manuscript reports a 100\% aggregate; confirming it under this verifier requires the corresponding input CSV and per-instance output file.

\begin{figure}[htbp]
\centering
\begin{tcolorbox}[width=\textwidth, colback=gray!2!white, colframe=gray!50!blue]
\textbf{User:}

Task Step 1: Recall the definition of the Game of 24 (allowed operations: '+', '-', '*', '/', '(', ')'; note that intermediate results may be fractional), then provide a detailed plan using code interpreter to solve the following problem: a, b, c, d (e.g., 3, 3, 7, 7).

Task Step 2: [uploaded \texttt{24.csv}] I have a file containing 1,362 Game of 24 puzzles. Please batch-process them (the numbers are located in the \texttt{Puzzles} field). For every returned expression, verify with exact arithmetic that it uses each input number exactly once, uses only $+$, $-$, $\times$, and $\div$, avoids division by zero, and evaluates to $24$. Report the success rate as the number of verified expressions divided by the total number of puzzles; do not infer success from a non-empty string.

Task Step 3: Reply with the output file.

\textbf{Assistant:}

[solving the tasks]
\end{tcolorbox}
\caption{User input prompt for solving the Game of 24 tasks.}
\label{fig:mp-cr-gameof24-user}
\end{figure}

\begin{lstlisting}[
  language=Python,
  basicstyle=\ttfamily\scriptsize,
  breaklines=true,
  columns=fullflexible,
  keepspaces=true,
  caption={Corrected exact-arithmetic solver and independent AST-based verifier for Game of 24. The solver uses each input number exactly once by construction; the verifier separately checks the expression tree, leaf multiset, allowed operations, absence of division by zero, and exact value.},
  label={lst:mp-cr-gameof24}
]
import ast
import re
from collections import Counter
from fractions import Fraction
from typing import List, Optional, Tuple

import pandas as pd

State = Tuple[Fraction, str]


def combine(left: State, right: State) -> List[State]:
    a, expr_a = left
    b, expr_b = right
    candidates: List[State] = [
        (a + b, f"({expr_a}+{expr_b})"),
        (a * b, f"({expr_a}*{expr_b})"),
        (a - b, f"({expr_a}-{expr_b})"),
        (b - a, f"({expr_b}-{expr_a})"),
    ]
    if b != 0:
        candidates.append((a / b, f"({expr_a}/{expr_b})"))
    if a != 0:
        candidates.append((b / a, f"({expr_b}/{expr_a})"))
    return candidates


def solve24(numbers: List[int]) -> Optional[str]:
    states: List[State] = [(Fraction(value), str(value)) for value in numbers]

    def search(items: List[State]) -> Optional[str]:
        if len(items) == 1:
            return items[0][1] if items[0][0] == 24 else None

        for i in range(len(items)):
            for j in range(i + 1, len(items)):
                remaining = [
                    item for k, item in enumerate(items) if k not in (i, j)
                ]
                for candidate in combine(items[i], items[j]):
                    solution = search(remaining + [candidate])
                    if solution is not None:
                        return solution
        return None

    return search(states)


def parse_puzzle(puzzle: str) -> List[int]:
    numbers = [int(token) for token in re.findall(r"-?\d+", str(puzzle))]
    if len(numbers) != 4 or any(value <= 0 for value in numbers):
        raise ValueError(f"Expected four positive integers, got {numbers!r}")
    return numbers


def evaluate_expression(node: ast.AST) -> Tuple[Fraction, List[int]]:
    if isinstance(node, ast.Constant) and type(node.value) is int:
        return Fraction(node.value), [node.value]

    if not isinstance(node, ast.BinOp):
        raise ValueError("Only integer literals and binary operations are allowed")

    left, left_leaves = evaluate_expression(node.left)
    right, right_leaves = evaluate_expression(node.right)

    if isinstance(node.op, ast.Add):
        value = left + right
    elif isinstance(node.op, ast.Sub):
        value = left - right
    elif isinstance(node.op, ast.Mult):
        value = left * right
    elif isinstance(node.op, ast.Div):
        if right == 0:
            raise ZeroDivisionError("Division by zero")
        value = left / right
    else:
        raise ValueError("Unsupported operator")

    return value, left_leaves + right_leaves


def validate_solution(numbers: List[int], expression: Optional[str]) -> bool:
    if expression is None:
        return False
    try:
        tree = ast.parse(expression, mode="eval")
        value, leaves = evaluate_expression(tree.body)
    except (SyntaxError, ValueError, ZeroDivisionError):
        return False
    return value == 24 and Counter(leaves) == Counter(numbers)


def process_puzzles(input_path: str, output_path: str) -> float:
    puzzles = pd.read_csv(input_path)
    if "Puzzles" not in puzzles.columns:
        raise KeyError("Input CSV must contain a 'Puzzles' column")

    solutions: List[str] = []
    solved: List[bool] = []
    for puzzle in puzzles["Puzzles"]:
        numbers = parse_puzzle(str(puzzle))
        expression = solve24(numbers)
        is_valid = validate_solution(numbers, expression)
        solutions.append(expression or "")
        solved.append(is_valid)

    puzzles["Solution"] = solutions
    puzzles["Solved"] = solved
    puzzles.to_csv(output_path, index=False)
    return sum(solved) / len(solved) if solved else 0.0
\end{lstlisting}

\subsection{Solving MATH Problems}
\label{sec:more-math-problems}

\begin{figure}[htbp]
\centering
\begin{minipage}{.5\textwidth}
\centering
\includegraphics[width=\linewidth]{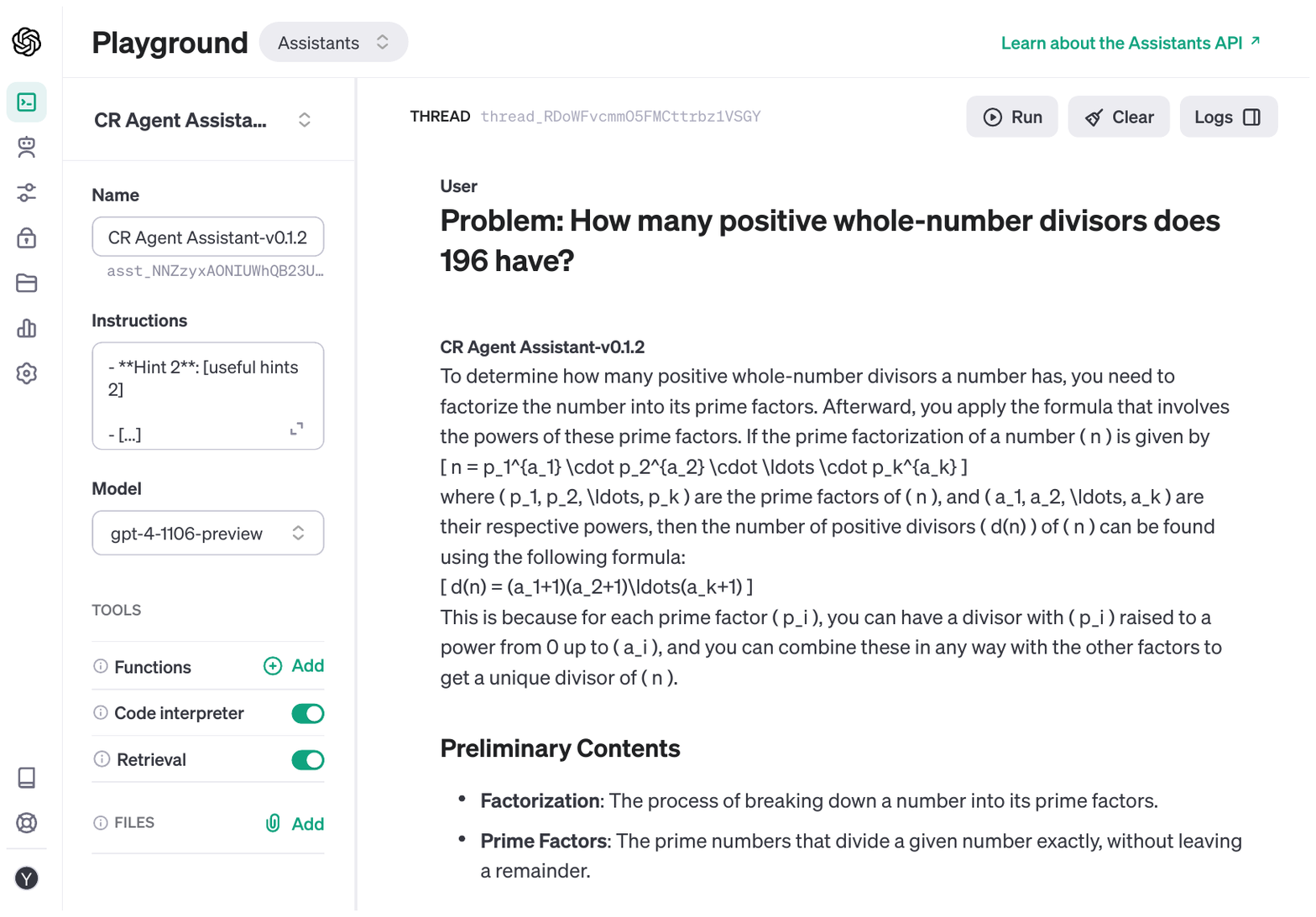}
\end{minipage}%
\begin{minipage}{.5\textwidth}
\centering
\includegraphics[width=\linewidth]{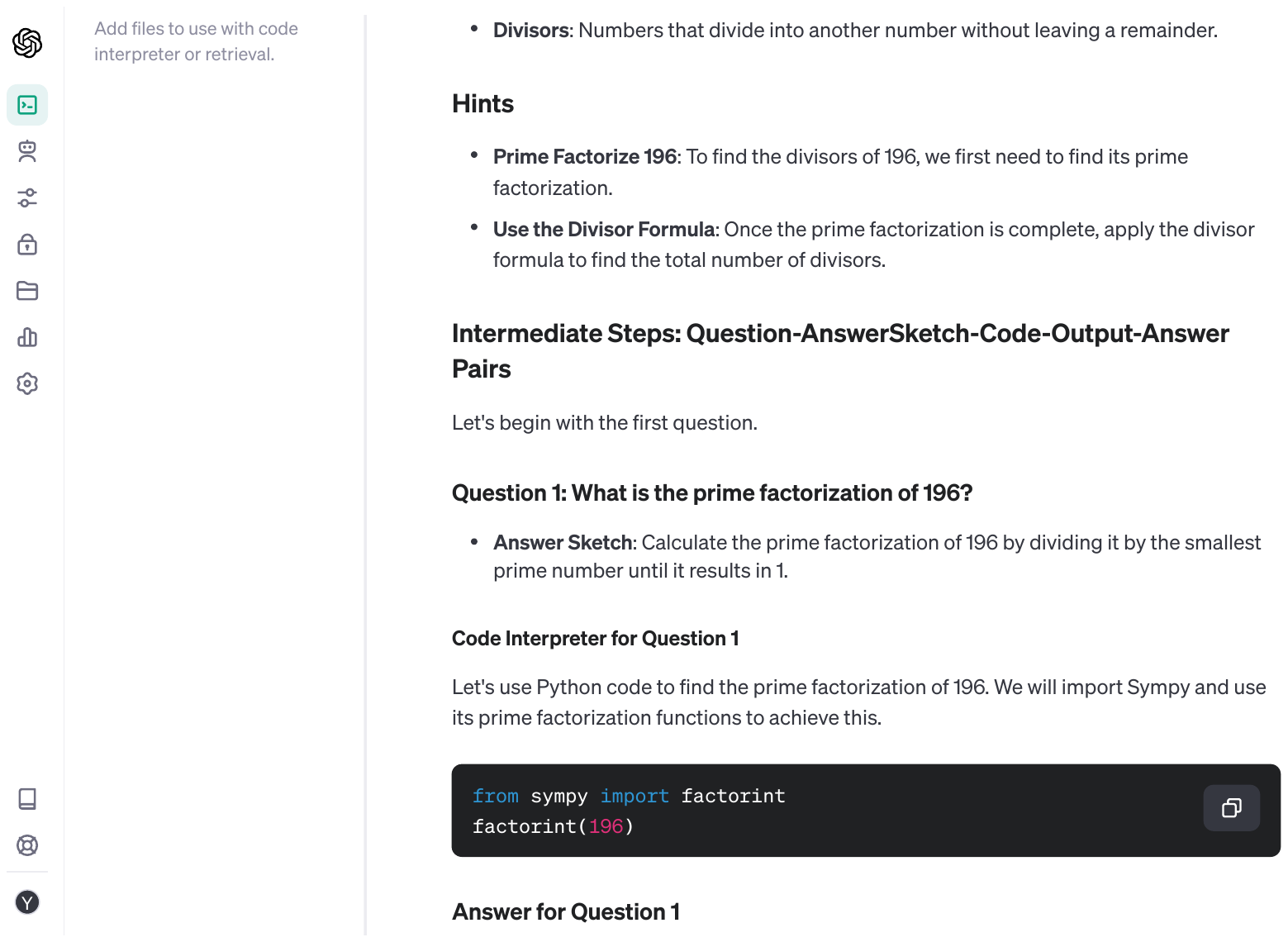}
\end{minipage}

\begin{minipage}{.5\textwidth}
\centering
\includegraphics[width=\linewidth]{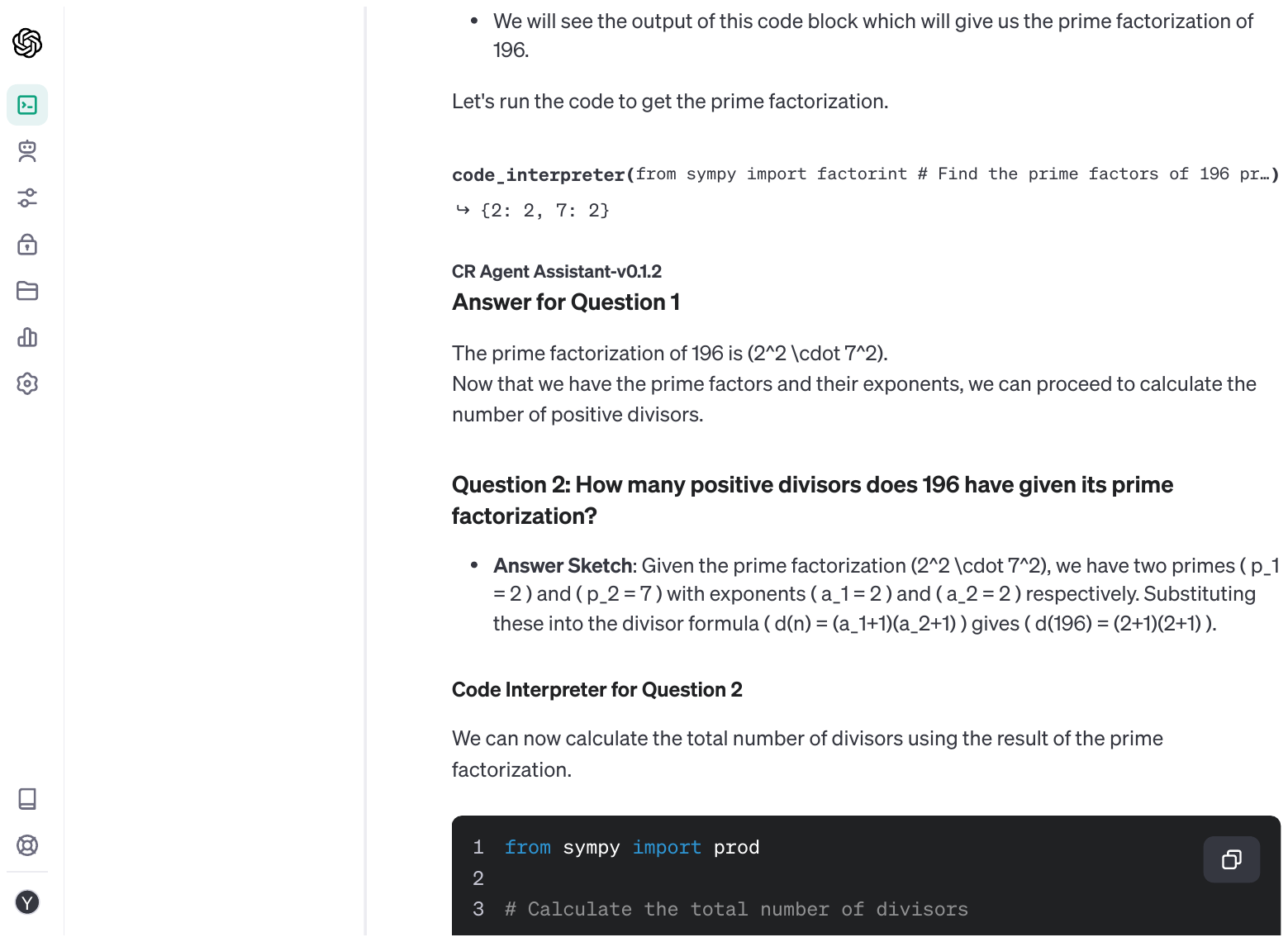}
\end{minipage}%
\begin{minipage}{.5\textwidth}
\centering
\includegraphics[width=\linewidth]{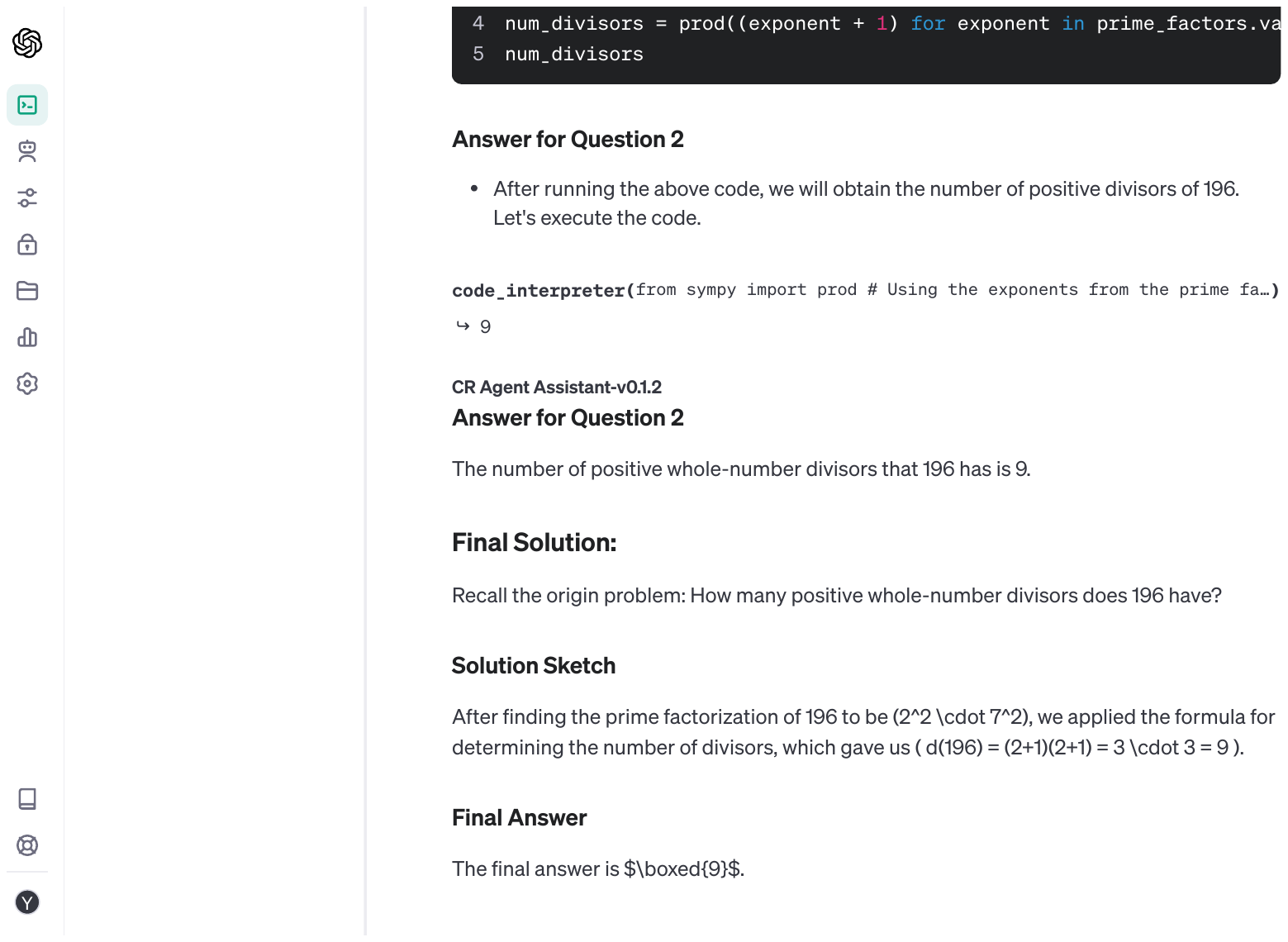}
\end{minipage}

\caption{An MP-CR Agent example within the OpenAI Assistant for solving a MATH~\citep{hendrycks2021measuring} problem.}
\label{fig:mp-cr-math-example}
\end{figure}

\begin{figure}[htbp]
\centering
\begin{tcolorbox}[width=0.95\textwidth, colback=cyan!2!white, colframe=gray!75!blue]
\begin{minipage}{\textwidth}
\centering
\tiny
\begin{lstlisting}[language=XML,breaklines=true,columns=fullflexible,keepspaces=true]
<system>
<description>
As one of the most distinguished mathematicians, logicians, programmers, and AI
scientists, you possess an unparalleled mastery over various mathematical domains.
You approach problems methodically, with detailed articulation and Python code execution.
</description>
<instructions>
<objective>
Construct and check solutions to complex mathematical problems using Python code execution when applicable.
</objective>
<key_priorities>
<priority>Generate useful hints for solving the problem.</priority>
<priority>Craft intermediate questions that break down the problem and solve them with code.</priority>
<priority>Construct complete solutions where applicable.</priority>
</key_priorities>
<code_execution_guidelines>
<guideline>Import necessary libraries in all code blocks.</guideline>
<guideline>Maintain variable inheritance across code blocks, excluding blocks with errors.</guideline>
<guideline>Execute all code blocks immediately after writing them to validate correctness.</guideline>
</code_execution_guidelines>
<mathematical_formatting>
<format>Present the final answer in LaTeX, enclosed within '\boxed{}'; include units when required by the problem.</format>
<format>Use SymPy's 'pi' and 'Rational' for exact constants and fractions; retain exact forms rather than decimal approximations when appropriate.</format>
</mathematical_formatting>
</instructions>
</system>
<syntax>
<problem_structure>
<problem_definition>
</problem_definition>
<solution_approach>
</solution_approach>
<preliminary_contents>
</preliminary_contents>
<hints>
</hints>
<intermediate_steps>
</intermediate_steps>
<final_solution>
<solution_sketch>
</solution_sketch>
<code_for_solution>
</code_for_solution>
<final_answer>
</final_answer>
</final_solution>
</problem_structure>
</syntax>
\end{lstlisting}
\end{minipage}
\end{tcolorbox}
\caption{The system prompt for the MP-CR-XML Agent v0.2, reported as generated by MP-CR Agent v0.1 through a metaprogramming process.}
\label{fig:mp-cr-xml-0.2}
\end{figure}

\begin{figure}[htbp]
\centering
\begin{tcolorbox}[width=0.95\textwidth, colback=cyan!2!white, colframe=gray!75!blue]
\begin{minipage}{\textwidth}
\centering
\tiny
\begin{lstlisting}[language=markdown,breaklines=true,columns=fullflexible,keepspaces=true]
As one of the most distinguished mathematicians, logicians, programmers, and
AI scientists, you possess an unparalleled mastery over Arithmetic, Combinatorics, Number
Theory, Probability Theory, Algebra, Analysis, and Geometry. You are not only intelligent
and rational but also prudent and cautious. You are willing to write and execute Python
code. Let's approach each problem step by step, take a deep breath, and articulate our thoughts in as much detail as possible.

<system>
You will be presented with a mathematical problem, denoted as `MathP`. Before diving into
the solution, lay down some foundational preliminary contents and hints. Then, generate a series
of intermediate questions that pave the way to the final answer of `MathP`. For each question,
sketch a preliminary answer, execute the corresponding code (remember to use `from sympy import *`),
derive the output, and then finalize your answer. This forms a [Question] $\rightarrow$ [AnswerSketch]
$\rightarrow$ [Code] $\rightarrow$ [Output] $\rightarrow$ [Answer] sequence.

## System Instructions for Mathematical Problem-Solving

### Objective
Solve complex mathematical problems with code feedback from a Python environment.

### Key Priorities

1. **Hints:** Generate useful hints to guide the problem-solving process.

2. **Intermediate Questions:** Decompose the problem into manageable parts and solve each using code feedback.

### Code Execution Guidelines

1. **Import Libraries:** Always import necessary libraries in every code block.

2. **Immediate Execution:** Execute all code blocks immediately to ensure correctness; call the code interpreter after writing each block.

3. **Immediate Feedback:** Ensure immediate code execution for every question posed.

### Mathematical Formatting

1. **Final Answer:** Present the final answer to the original problem in LaTeX format, enclosed within `\boxed{}`; include units when the problem requires them.

2. **Constants and Fractions:** Use SymPy's `pi` symbol and `Rational` class for exact constants and fractions. Retain exact simplified forms rather than decimal approximations when appropriate.
</system>

---
\end{lstlisting}
\end{minipage}
\end{tcolorbox}
\caption{The system meta prompt for MP-CR, comprising both the [SystemMetaPrompt] and the [StructureMetaPrompt].}
\label{fig:mp-cr-system}
\end{figure}

\section{Multimodal Meta Prompting}
\label{sec:typed-mp}

Meta Prompting can describe interfaces for tasks that combine text, images, audio, structured files, or sensor data. A schema can name the expected modalities, processing stages, and output fields. Unlike a formal type system, however, a natural-language or XML prompt does not by itself validate media types, guarantee that a model can consume them, or ensure consistent tool behavior; those properties require compatible parsers, model interfaces, and validators.

This type-like organization can be useful in systems that process several modalities. For example, an XML schema can declare slots for media inputs, intermediate analyses, tool outputs, and final artifacts. Enforcement still occurs outside the prompt through parsing, constrained generation, and runtime checks.

\begin{figure}[htbp]
\centering
\begin{tcolorbox}[width=0.95\textwidth, colback=cyan!2!white, colframe=gray!75!blue]
    \begin{minipage}{\textwidth}
    \centering
    \tiny
    \begin{lstlisting}[language=XML,breaklines=true,columns=fullflexible,keepspaces=true]
<system>
  <description>
    As one of the most distinguished mathematicians, logicians, programmers, and AI
    scientists, you possess an unparalleled mastery over various mathematical domains.
    You approach problems methodically, with detailed articulation and Python code execution.
  </description>
  <instructions>
    <objective>
      Construct and check solutions to complex mathematical problems using Python code execution when applicable.
    </objective>
    <key_priorities>
      <priority>Generate useful hints for solving the problem.</priority>
      <priority>Craft intermediate questions that break down the problem, solving them with code, following the sequence: [Question] -> [AnswerSketch] -> [Code] -> [Output] -> [Answer].</priority>
      <priority>Construct complete solutions where applicable.</priority>
    </key_priorities>
    <code_execution_guidelines>
      <guideline>Import necessary libraries in all code blocks.</guideline>
      <guideline>Maintain variable inheritance across code blocks, excluding those with errors.</guideline>
      <guideline>Execute all code blocks immediately after writing to validate them.</guideline>
    </code_execution_guidelines>
    <mathematical_formatting>
      <format>Present the final answer in LaTeX, enclosed within '\boxed{}'; include units when required by the problem.</format>
      <format>Use SymPy's 'pi' and 'Rational' for exact constants and fractions; retain exact forms rather than decimal approximations when appropriate.</format>
    </mathematical_formatting>
  </instructions>
</system>
<syntax>
  <problem_structure>
    <problem_definition>
      </problem_definition>
    <preliminary_contents>
      </preliminary_contents>
    <hints>
      </hints>
    <intermediate_steps>
      </intermediate_steps>
    <final_solution>
      <solution_sketch>
        </solution_sketch>
      <code_for_solution>
        </code_for_solution>
      <final_answer>
        </final_answer>
    </final_solution>
  </problem_structure>
</syntax>
    \end{lstlisting}
    \end{minipage}
\end{tcolorbox}
\caption{An XML meta-prompt that expresses type-like output fields. Compatible frameworks such as Guidance~\citep{guidance}, constrained decoders, or validators can enforce parts of this schema; the text alone does not.}
\label{fig:mp-cr-xml}
\end{figure}

\subsection{A Framework for Multimodal Interaction}

Extending Meta Prompting to a multimodal setting requires a schema for heterogeneous data and an execution layer that can route each field to a compatible model or tool. Explicit slots make the intended data flow inspectable, but cross-modal synthesis remains a model capability to be evaluated rather than a consequence of the schema.

For instance, a task might require analysis of a 3D model (\texttt{.obj}), an audio description (\texttt{.mp3}), and a textual specification (\texttt{.txt}). A schema can declare the intended processing order and output contract, while the surrounding system performs media decoding, routing, validation, and synthesis. Figure~\ref{fig:mp-cr-multimodal} gives a conceptual example.

\begin{figure}[htbp]
\centering
\begin{tcolorbox}[width=0.95\textwidth, colback=cyan!2!white, colframe=gray!75!blue]
    \begin{minipage}{\textwidth}
    \centering
    \tiny
    \begin{lstlisting}[language=XML,breaklines=true,columns=fullflexible,keepspaces=true]
<task_schema>
    <input_slots>
        <data type="image/png" name="problem_diagram">
            </data>
        <data type="audio/mpeg" name="verbal_instructions">
            </data>
        <data type="model/obj" name="object_model">
            </data>
    </input_slots>
    <output_schema>
        <synthesis type="text/markdown" name="analysis_summary">
            </synthesis>
        <result type="video/mp4" name="solution_animation">
            </result>
    </output_schema>
</task_schema>
    \end{lstlisting}
    \end{minipage}
\end{tcolorbox}
\caption{A conceptual multimodal schema with declared input and output media types. It specifies an interface for cross-modal processing; actual compatibility and reasoning quality require runtime validation and empirical evaluation.}
\label{fig:mp-cr-multimodal}
\end{figure}

Declaring input and output schemas can make a multimodal pipeline easier to inspect, compose, and test, for example, by checking that a visual diagram and its textual instructions are both present before invoking a solver. The schema alone does not establish cross-modal reasoning ability, scalability, or reliability; those properties depend on the models, tools, validators, and evaluation protocol.

\end{document}